\documentclass[conference]{IEEEtran}
\IEEEoverridecommandlockouts
\usepackage{cite}
\usepackage{amsmath,epsfig,algorithm}
\usepackage{amsmath,amssymb,amsfonts}
\usepackage{algpseudocode}
\usepackage{graphicx}
\usepackage{textcomp}
\usepackage{xcolor}
\usepackage{todonotes}
\usepackage{booktabs}
\usepackage{comment}
\newcommand{\ra}[1]{\renewcommand{\arraystretch}{#1}}

\DeclareMathOperator*{\argmin}{arg\,min}
\def\BibTeX{{\rm B\kern-.05em{\sc i\kern-.025em b}\kern-.08em
    T\kern-.1667em\lower.7ex\hbox{E}\kern-.125emX}}
\begin{document}

\title{DPGOMI: Differentially Private Data Publishing with Gaussian Optimized Model Inversion 
\thanks{
Accepted by 2023 IEEE International Conference on Image Processing (ICIP) workshop on privacy attacks in computer vision, Kuala Lumpur, Malaysia, 2023,

Research reported in this project was supported by the National Institutes of Health, United States of America under award number R01MH121344 and the Child Family Endowed Professorship.
This work was supported in part by Oracle Cloud credits and related resources provided by the Oracle for Research program.

© 2023 IEEE. Personal use of this material is permitted. Permission from IEEE must be obtained for all other uses, in any current or future media, including reprinting/republishing this material for advertising or promotional purposes, creating new collective works, for resale or redistribution to servers or lists, or reuse of any copyrighted component of this work in other works.}
}

\author{Dongjie Chen$^{1}$~~~
Sen-ching S. Cheung$^{1,2}$~~~
Chen-Nee Chuah$^1$~~~ \\
$^1$ECE Department,
	University of California,
	Davis, CA \\
 $^2$ECE Department,
	University of Kentucky,
	Lexington, KY
}

\maketitle

\begin{abstract}
High-dimensional data are widely used in the era of deep learning with numerous applications. However, certain data which has sensitive information are not allowed to be shared without privacy protection. In this paper, we propose a novel differentially private data releasing method called Differentially Private Data Publishing with Gaussian Optimized Model Inversion (DPGOMI) to address this issue. Our approach involves mapping private data to the latent space using a public generator, followed by a lower-dimensional DP-GAN with better convergence properties. We evaluate the performance of DPGOMI on standard datasets CIFAR10 and SVHN. Our results show that DPGOMI outperforms the standard DP-GAN method in terms of Inception Score, Fréchet Inception Distance, and classification performance, while providing the same level of privacy. Our proposed approach offers a promising solution for protecting sensitive data in GAN training while maintaining high-quality results.
\end{abstract}

\begin{IEEEkeywords}
Generative adversarial networks, differential privacy, Gaussian optimized model inversion
\end{IEEEkeywords}

\section{Introduction}

Generative AI is one of the major technological breakthroughs in the 21st century. All major generative AI models are trained on a large amount of diverse data that might contain personally identifiable information (PII). To address the privacy concerns associated with generative models, the standard approach is to use differentially private (DP) stochastic gradient descent, which adds controlled noise to the gradients during training~\cite{abadi2016deep}. In the presence of such noise, the quality of the output synthetic samples are adversely affected and the training of the network may not converge~\cite{bu2021convergence}. 

In this paper, we propose a novel DP generative model using Gaussian Optimized Model Inversion (DPGOMI) method, where private data is first mapped to the latent space of a public generator, followed by a low-dimensional DP-GAN to model the distribution of the private latent vectors. This is an extension of our prior work in \cite{chen2021differentially}.
Different from the scheme in \cite{chen2021differentially} that restricts the model inversion process with a clipping method, our proposed scheme improves the model inversion process by introducing a Gaussian modulated cost function, aiming to further improve the quality of synthetic private data. 

We have evaluated the performance of DPGOMI on two standard datasets: CIFAR10~\cite{krizhevsky2009learning} and SVHN~\cite{netzer2011reading}. Our experimental results show that DPGOMI outperforms the state-of-the-art (SOTA) methods in terms of Inception Score, Fréchet Inception Distance, and classification precision while providing the same level of privacy guarantee. Particularly, DPGOMI has up to 47\% improvement in classification precision for downstream classification tasks.


\section{Related Work} 
The standard method to assure privacy in machine learning involves the implementation of different differential privacy (DP) algorithms, typically based on the Gaussian mechanism~\cite{dwork2014algorithmic}. Techniques such as Differentially Private Stochastic Gradient Descent (DPSGD), and Private Aggregation of Teacher Ensembles (PATE)~\cite{papernotscalable} have been widely applied to the training process of generative models. 

DPSGD~\cite{abadi2016deep} privatizes a deep learning model by adding Gaussian noise to the aggregated gradients for SGD-based optimizers. 
When DPSGD is applied to a GAN, noise is only injected during the training of the discriminator, but not the training of the generator, which is not exposed to the private training data. The amount of injected noise is proportional to the number of parameters in the discriminator, and the privacy budget puts a limit on the number of training steps~\cite{chen2021differentially, harder2023pre}. As a result, it is challenging to scale up the input dimension of a GAN trained with DPSGD as the excessive amount of DP noise and the longer training process needed to build a larger model often leads to convergence issues or mode collapse~\cite{fan2020survey}.

To address the scalability issues of DPSGD on GANs, GS-WGAN~\cite{chen2020gs} proposed a gradient-sanitized approach with the Wasserstein objective as an alternative to the gradient clipping scheme used in DPSGD. This approach provides a better estimate on the sensitivity values used in determining the amount of DP noise needed. 
GS-WGAN also employed multiple discriminator networks trained on disjoint parts of the dataset to conserve the privacy budget through sub-sampling. A similar strategy was used in the PATE-GAN~\cite{jordon2018pate}, which deployed a single 'student' discriminator to be trained in DP fashion by a set of 'teacher' discriminators, each assigned to a distinct subset of the training data. The partition of training data, however, limits the usefulness of these approaches, especially in the typical scenarios when the training data is scarce.


Another strategy to improve the utilization of the privacy budget is to form a privacy barrier between the generator and the discriminator~\cite{long2021g}. By assuming that only the generator will be released to the public, G-PATE~\cite{long2021g} replaced the discriminator in a GAN with a private PATE classifier and passes the discretized gradients from the classifier to the generator. A similar approach was also adopted in GS-WGAN~\cite{chen2020gs}. However, their implementations rely on the use of conditional GAN, which limits its application to only fully-labeled datasets. 

The linkage between the discriminator and the generator can be further severed by training the generator using public data only. DPMI~\cite{chen2021differentially} proposed to learn a low dimensional DP latent space to model the distribution of private latent vectors generated from the private training data by applying model inversion (MI) on a public generator. The convergence was easier to achieve in the low-dimensional DP latent space. However, during training, the clipping method used in the MI process could result in low-quality synthetic images. An alternate line of works utilized Maximum Mean Discrepancy (MMD) between private target data and a generator's distribution based on random features or perceptual features learned from a public dataset~\cite{santos2019learning, harder2021dp, harder2023pre}. In this vein, DP-MERF~\cite{harder2021dp} and DP-MEPF~\cite{harder2023pre} introduced differentially private data generation methodologies through mean embeddings of features without the need of adding DP noise in every step of the training.

\section{Preliminaries}
\subsection{Differential Privacy}
A randomized mechanism, denoted as $M$, acting on a database $D$, is termed differentially private if the output of any potential query remains substantially unchanged when replacing $D$ with a neighboring database $D'$, which deviates from $D$ by no more than one data entry. This concept of a differentially private mechanism can be formally defined as follows:
\newtheorem{mydef}{Definition}
\begin{mydef}
(Differential Privacy) A randomized mechanism $M$ is $(\epsilon,\delta)$-differential privacy if any output set $S$ and any neighboring databases $D$ and $D'$ satisfy the followings:
\begin{equation}
\mathrm{P}(\mathcal{M}(D) \in S) \leq e^{\epsilon} \cdot \mathrm{P}\left(\mathcal{M}\left(D^{\prime}\right) \in S\right)+\delta
\end{equation}
\end{mydef}
For a $\delta$ value of zero, a smaller $\epsilon$ produces a more stringent privacy constraint, with an $\epsilon$ value of zero representing perfect privacy. The introduction of a small, positive $\delta$ value permits a minimal probability of failure, offering the benefit of more adaptable mechanism designs. Most DP randomized mechanisms used in machine learning are based on additive noise: Laplacian mechanism adds Laplacian noise the output to achieve $(\epsilon,0)$-DP while Gaussian mechanism adds Gaussian noise to achieve $(\epsilon,\delta)$-DP~\cite{dwork2014algorithmic}. 

A notable attribute of differential privacy mechanisms is that any post-processing conducted on a differentially private mechanism inherently preserves its differential privacy~\cite{dwork2014algorithmic}. This can be formally stated as follows:
\newtheorem{mydef2}{Theorem} 
\begin{mydef2}
(Post-processing) Given an arbitrary mapping $f: R \rightarrow R^{\prime}$ and an $(\epsilon,\delta)$-differentially private mechanism $\mathcal{M}: D \rightarrow R$, $f \circ \mathcal{M}: D \rightarrow R^{\prime}$ is $(\epsilon,\delta)$-differentially private. \label{thm:post}
\end{mydef2}

Note that the post-processing steps no longer have access to the private data. Typical deep network training, on the other hand, requires repeated exposure to the private data. In order to account for culminated privacy loss, an accountant scheme such as the Rényi Differential Privacy accountant~\cite{mironov2017renyi} should be used to bound the privacy loss within the total privacy budget $\epsilon$ while accounting for the composition of multiple queries.





\subsection{GAN}
A conventional Generative Adversarial Network (GAN) involves two components: a generator $G(z) \in X$ and a discriminator $C(x) \in {0,1}$. The generator employs a latent vector $z \in \mathbb{R}^d$ as a means to transform this latent representation into the target image space $X$. In contrast, the discriminator assesses whether an input image $x \in X$ is real (1) or synthetic (0). The latent vector's distribution, $P_Z$, is generally established as a separable $d$-dimensional Gaussian distribution. Provided that the real data is derived from a distribution $P_X$, the training objective of a GAN is to determine the optimal $G$ and $C$ by engaging in a two-player min-max game with the aim of solving the following optimization problem:
\begin{eqnarray}
\min_{G} \max_{C} \mathbb{E}_{x \sim P_X}[\log(C(x))] 
+\mathbb{E}_{z \sim P_Z}[\log(1-C(G(z)))] 
\end{eqnarray}
Wasserstein GAN (WGAN), introduces a significant enhancement to the training stability of the original GAN by leveraging the Wasserstein distance between the distributions of latent vectors and real images~\cite{arjovsky2017wasserstein}. Given a parametrized family of $K$-Lipschitz functions denoted as $\left\{f_{w}(x)\right\}_{w\in W}$, this optimization problem can be approximately tackled through the subsequent value function:
\begin{eqnarray}
\min_{G} \max_{w \in W} \mathbb{E}_{x \sim P_X}\left[f_{w}(x)\right]-\mathbb{E}_{z \sim P_Z}\left[f_{w}(G(z))\right]
\end{eqnarray}

\subsection{Model Inversion of GANs}
\label{subsec:MI}

Using a pre-trained public generator $G_{p}$, which transforms a random $d$-dimensional latent vector $z \sim P_{Z} = \mathcal{N}^d(0,I)$ into a synthetic image $G_{p}(z)$, model inversion of GANs~\cite{chen2021differentially} aim to find, for each private image $x_{s} \in D_{s}$, the associated latent vector $z_{s}$ that minimizes the mean square difference between $x_{s}$ and $G_{p}(z_{s})$. Specifically, model inversion addresses the following optimization problem:

\begin{equation} 
\begin{aligned}
z_s = \argmin_{z}      \quad & ||G_{p}(z) - x_{s}||^{2}\\
\textrm{s.t.} \quad & \mbox{$P_{Z}(z) \geq P_{Z}(z_0)$ with $z_0 \sim P_{Z}$}  \label{eq:MI} \\
\end{aligned}
\end{equation}

Model inversion starts by first drawing a random sample $z_0$ from $P_{Z}$ and the constraint in (\ref{eq:MI}) is to ensure that the quality of $G_p(z_s)$ to be comparable to that of $G_p(z_0)$. This convex optimization can be easily solved via stochastic gradient ascent procedure by projecting the search trajectory back onto the convex constraint~\cite{shalev2014understanding}.

\section{Proposed Method}
\label{sec:method}

\begin{figure}[h!]
\centering
\includegraphics[width=\linewidth]{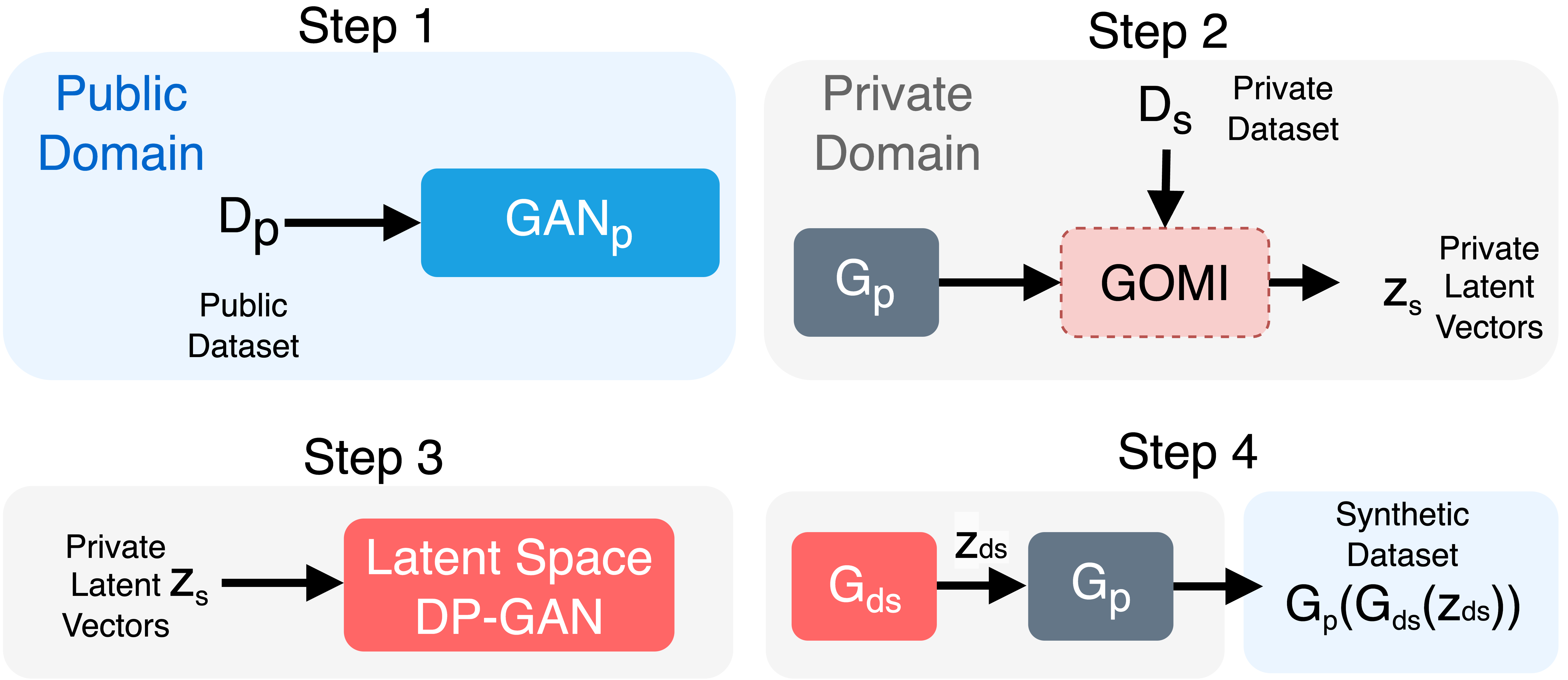}
\caption{Proposed DPGOMI Framework.}
\label{fig:overall_frame}
\centering
\end{figure}

In this section, we present a new differentially private data-releasing method named DPGOMI. An overview of the method is shown in Figure~\ref{fig:overall_frame}. First, instead of using the private data to train a differentially private GAN, we use public data $D_p$ similar to the private dataset $D_s$ to train a public GAN. Since the $D_p$ is publicly accessible, no DP training is needed for the public GAN. 
Next, an improved version of MI introduced in Section \ref{subsec:MI} called Gaussian Optimized Model Inversion (GOMI) is used to obtain the private latent vectors form the latent space of the public generator $G_p$. In step three, a low-dimensional differentially private GAN is trained to generate latent vectors that mimic the distribution of the private latent vectors. A synthetic dataset can be released in the final step by sampling the latent vector space and feeding them through the two generators: $G_{ds}$ from the DP latent space and the public generator $G_{p}$.

To motivate the need of GOMI, we recall that the goal of the MI process in Section \ref{subsec:MI} is to identify the latent vector $\hat{z}$ so that the synthetic image $G_p(\hat{z})$ is as close to the target private image $x_s$ as possible. To avoid the gradient descent search process from drifting too far off the prescribed standard Gaussian distribution, $P_{Z}=\mathcal{N}^d(0,I)$, the algorithm in Section \ref{subsec:MI} uses a constraint $P_{Z}(z) \geq P_{Z}(z_0)$ with $z_0 \sim P_{Z}$. The random $z_0$, however, does not provide a deterministic guarantee of the quality of the synthetic image. The truncation trick in \cite{marchesi2017megapixel} used a truncated Gaussian distribution during the GAN inference process to provide such a deterministic guarantee. But the truncated Gaussian function is not entirely differentiable so it impacts the stochastic gradient search process in MI.
We introduce a differentiable cost function that penalizes the deviation from the latent distribution $P_Z(z)$. We call this new approach Gaussian Optimized Model Inversion (GOMI) which can be described as follows:
\begin{equation} 
\begin{aligned}
z_s = \argmin_{z}      \quad & ||G_{p}(z) - x_{s}||^{2} / P_Z(z)
\label{eq:GMI} \\
\end{aligned}
\end{equation}
where $P_Z(z)$ represents the value of the standard Gaussian function at latent vector $z$. GOMI exposes the model inversion process to a broader latent space while ensuring high image quality. The detailed algorithm to implement GOMI with the Adam optimizer is shown in Algorithm~\ref{alg:gomi-adam}. 

\begin{algorithm}[h!]
\small
\caption{Gaussian Optimized Model Inversion (GOMI) using the Adam optimizer}
\label{alg:gomi-adam}
\begin{algorithmic}[1]
\State \textbf{Input:} Generator $G_p$, private image $x_s$, Adam parameters $\beta_1, \beta_2$, learning rate $\alpha$, \# iterations $N$, standard Gaussian $P_Z$, small tolerance to prevent division by zero $e$
\State Initialize $z \sim P_{Z}$
\State Initialize moment vectors $\mathbf{m}, \mathbf{v} \leftarrow \mathbf{0}$ 
\For{$i = 1$ to $N$}
\State $\mathbf{g} \leftarrow \nabla_{z} (||G_{p}(z) - x_{s}||^{2} / P_Z(z))$
\State Update $\mathbf{m} \leftarrow \beta_1 \mathbf{m} + (1 - \beta_1) \mathbf{g}$
\State Update $\mathbf{v} \leftarrow \beta_2 \mathbf{v} + (1 - \beta_2)\mathbf{g}$
\State Bias correction: $\hat{\mathbf{m}} \leftarrow \mathbf{m} / (1 - \beta_1^i)$
\State Bias correction: $\hat{\mathbf{v}} \leftarrow \mathbf{v} / (1 - \beta_2^i)$
\State $z \leftarrow z - \alpha \hat{\mathbf{m}} / (\sqrt{\hat{\mathbf{v}}} + e)$
\EndFor
\State \textbf{Output:} The optimized latent vector $z_t$
\end{algorithmic}
\end{algorithm}

\begin{table*}[h!]
\centering
\ra{1.0}
\caption{FID values for different methods and $\epsilon$ values for CIFAR10 and SVHN ($\delta=10^{-5}$)}
\label{tab:all_fid}
\begin{tabular}{@{}lrrrrrrrrrrr@{}}
\toprule
\multicolumn{3}{}{} & CIFAR10 &\multicolumn{5}{}{} & SVHN \\ \cmidrule{2-6} \cmidrule{8-12}
Method & $\epsilon = 1$ & $\epsilon = 5$ & $\epsilon = 10$ & $\epsilon = 20$ & $\epsilon = 50$ & & $\epsilon = 1$ & $\epsilon = 5$ & $\epsilon = 10$ & $\epsilon = 20$ & $\epsilon = 50$ \\
\midrule
DP-GAN~\cite{xie2018differentially} & 323.27 & 329.80 & 336.21 & 255.29 & 247.40 & & 306.54 & 295.11 & 297.73 & 290.70 & 253.32 \\
DP-MERF~\cite{harder2021dp} & 331.28 & 325.04 & 324.78 & 312.54 & 307.72 & & 344.58 & 338.22 & 327.84 & 320.06 & 310.43 \\
G-PATE~\cite{long2021g} (cond.) & 444.56 & 439.19 & 347.86 & 309.03 & 309.03 & & 461.00 & 416.02 & 461.08 & 402.51 & 400.57 \\
GS-WGAN~\cite{chen2020gs} (cond.) & 354.46 & 275.70 & 233.30 & 223.62 & 223.62 & & 302.13 & 158.38 & 162.19 & 161.48 & 119.4 \\
DP-MEPF~\cite{harder2023pre} ($\phi_1, \phi_2$) & 175.50 & 166.88 & 151.48 & 152.24 & 152.91 & & 113.54 & 95.91 & 120.22 & 115.90 & 87.15 \\
DP-MEPF~\cite{harder2023pre} ($\phi_1$) & 132.57 & 128.92 & 124.01 & 111.99 & 104.98 & & 101.05 & 93.16 & 82.60 & 81.76 & 78.69 \\  
DPMI~\cite{chen2021differentially} & 130.61 & 121.67 & 108.06 & 104.47 & 97.68 & & 72.27 & 83.96 & 72.67 & 67.91 & 63.62 \\
DPGOMI  & \bf{127.67} & \bf{95.54} & \bf{94.45} & \bf{93.67} & \bf{93.14} & & \bf{70.13} & \bf{67.47} & \bf{65.47} & \bf{55.64} & \bf{53.88} \\
DPGOMI ($\epsilon$=$\infty$) & 81.18 & 81.18 & 81.18 & 81.18 & 81.18 & & 42.64 & 42.64 & 42.64 & 42.64 & 42.64 \\
\bottomrule
\end{tabular}
\end{table*}

\begin{table}[ht]
\centering
\ra{1.0}
\caption{Inception Score and Downstream Classification Precision comparison on $\epsilon=10$} \label{tab:IS_classification}
\begin{tabular}{@{}lccccc@{}}
\toprule
& \multicolumn{2}{c}{Inception Score} & &\multicolumn{2}{c}{Classification} \\ \cmidrule{2-3} \cmidrule{5-6} 
& CIFAR10 & SVHN & & CIFAR10 & SVHN \\ \cmidrule{2-3} \cmidrule{5-6} 
DP-GAN~\cite{xie2018differentially} & 1.67 & 1.73 & &  0.28 & 0.32 \\
G-PATE~\cite{long2021g} (cond.) & 1.29 & 1.46 & & 0.32 & 0.45 \\
GS-WGAN~\cite{chen2020gs} (cond.) & 1.88 & 1.63 & & 0.31 & 0.35 \\
DP-MERP~\cite{harder2021dp} & 2.95 & 2.39 & & 0.35 & 0.53 \\
DP-MEPF~\cite{harder2023pre} ($\phi_1$, $\phi_2$) & 3.05 & 2.44 & & 0.67 & 0.67 \\
DP-MEPF~\cite{harder2023pre} ($\phi_1$) & 2.97 & 2.61 & & 0.71 & 0.77 \\
DPMI~\cite{chen2021differentially} & 4.46 & 2.07 & & 0.67 & 0.69 \\
DPGOMI & \bf{4.74} & \bf{2.59} & & \bf{0.73} & \bf{0.79} \\
DPGOMI ($\epsilon$=$\infty$) & 4.80 & 2.76 & & 0.86 & 0.92 \\
\bottomrule
\end{tabular}
\end{table}

\section{Experiments}
\label{sec:experiments}

In this section, we analyze the performance of our proposed DPGOMI method against the state-of-the-art differentially private data release techniques at various levels of privacy. Additionally, we conduct an in-depth ablation study of the GOMI component. To evaluate the released data, we assess the quality of the synthetic images using standard metrics for image quality and diversity. 

\subsection{Dataset Partition}

Following~\cite{chen2021differentially}, we separate the dataset into public and private domains. We randomly allocate one-third of the training set to $D_l$, which we use to train the labeling classifier. We then evenly split the rest of the training set into $D_p$ and $D_s$, assigning half the classes to each. The public GAN is trained using $D_p$, and the GOMI process uses $D_s$. We only use half of the testing set with private labels. 

\subsection{Datasets and Evaluation Metrics}
\label{ssec:expermentsetting}
Two public datasets are used in our experiments:
\begin{enumerate}
    \item The CIFAR10 dataset~\cite{krizhevsky2009learning} contains 50,000 training and 10,000 testing images, categorized into ten classes. Each image is of size 32$\times$32$\times$3. For our experiments, the classes automobile, bird, cat, deer, and dog classes are randomly to the public set $D_p$, while frog, horse, ship, truck, and airplane classes are used for $D_s$.
    \item SVHN~\cite{netzer2011reading} is a door-sign digit image dataset, containing 73,257 training and 26,032 testing images of size 32$\times$32$\times$3. The public set $D_p$ contains digits 1, 5, 7, 8 and 9 and $D_s$ contains digits 0, 2, 3, 4 and 6.
\end{enumerate}

For performance measurements, we use classification Inception Score (IS) and Fréchet Inception Distance (FID) based on InceptionV3 to measure both synthetic image quality and diversity~\cite{borji2021pros}. We also evaluate the quality of the images using CNN based downstream classification tasks.


\subsection{Comparison with SOTA DP generative models}
In this Section, we compare the DPGOMI scheme with six SOTA DP generative models~\cite{xie2018differentially, long2021g, chen2020gs, harder2021dp, harder2023pre, chen2021differentially}. Since only one-third of the training set is used as $D_s$, we modify the open-source codes of these papers to adapt our data partition settings and obtain their performance for comparison. 
For G-PATE and GS-WGAN, we follow the settings in the original papers and use conditional generation. We use unconditional generation on all the other methods, including ours. $D_p$ is used in DP-MEPF for public training.

We first evaluate the quality of the synthetic data using FID. A lower FID value indicates higher image quality. Table~\ref{tab:all_fid} shows the trade-off between FID and the privacy level $\epsilon$ for the two datasets. DPGOMI outperforms all the existing methods across all the $\epsilon$ for both CIFAR10 and SVHN. 

We then compare the IS and downstream classification precision with $\epsilon$=10. A larger IS indicates better quality and diversity of the synthetic images. Table \ref{tab:IS_classification} shows that DPGOMI has a higher IS than all the other models when they guarantee the same amount of privacy for both CIFAR10 and SVHN. DPGOMI also has a 2\% to 45\% improvement on CIFAR10 and a 2\% to 47\% improvement on SVHN for downstream classification tasks. 
Keep in mind that our private dataset include images from only half of the classes. This makes our measurements lower than the best available results. 


\subsection{Ablation study on Gaussian Optimized Model Inversion}
To show the effectiveness of Gaussian Optimized Model Inversion, we evaluate GOMI under an ablation study by comparing it with the MI method from~\cite{chen2021differentially}.

\begin{figure}[hbt!]
\begin{minipage}[b]{0.475\linewidth}
  \centering
  \includegraphics[width=\linewidth]{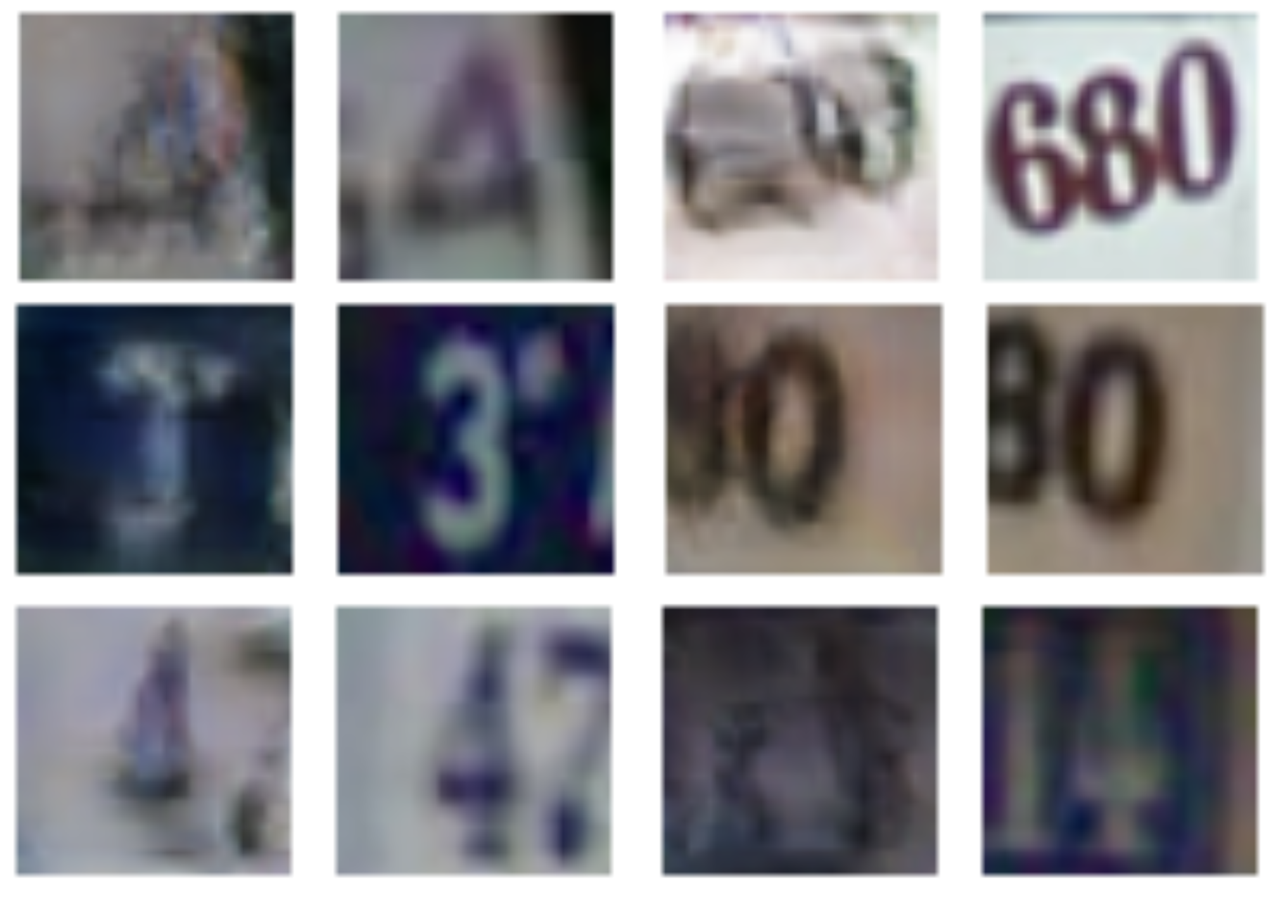}
  \centerline{(a)}
\end{minipage}\hfill
\begin{minipage}[b]{0.475\linewidth}
  \centering
  \includegraphics[width=\linewidth]{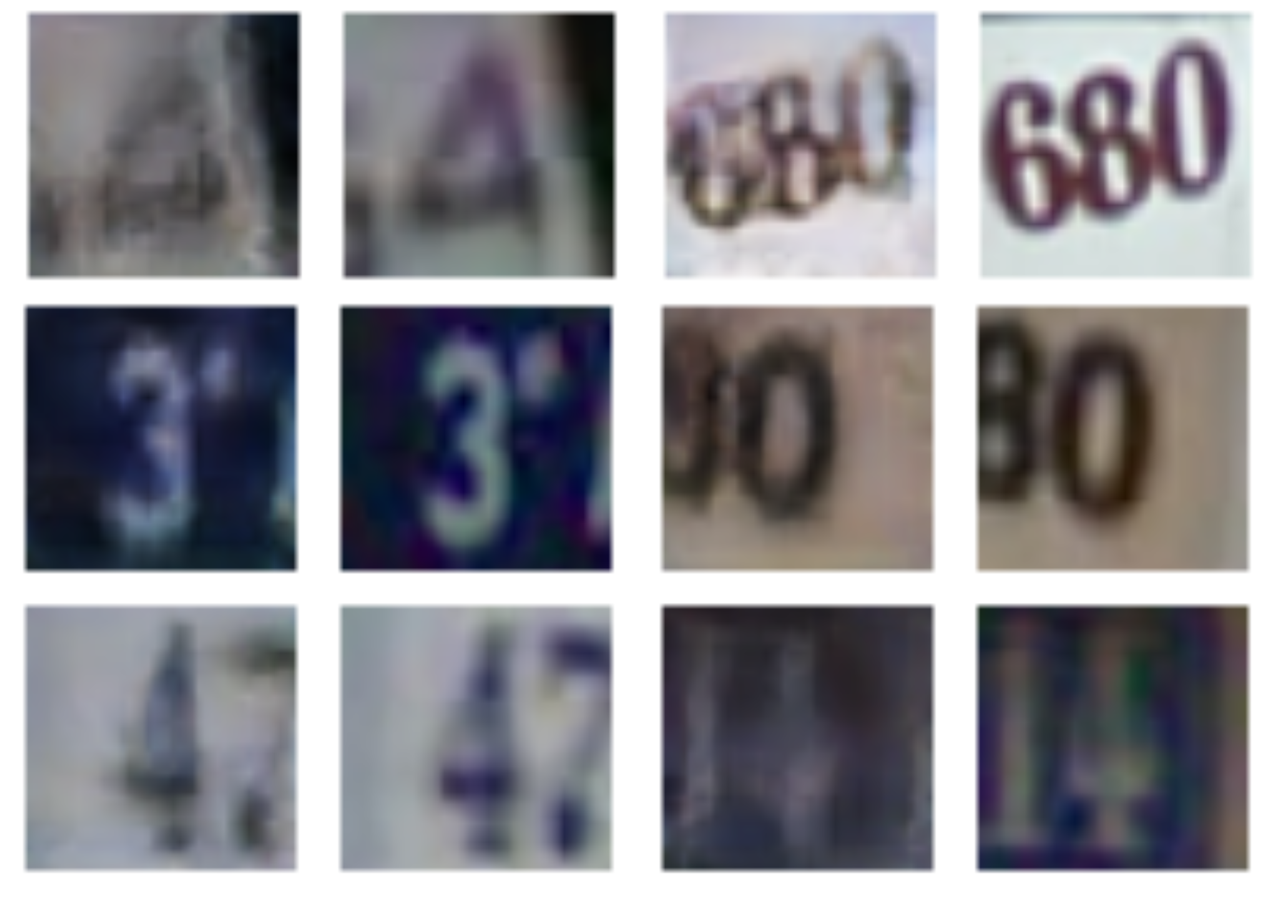}
  \centerline{(b)}
\end{minipage}
\caption{ Side-by-side comparisons with even columns for real images and odd columns for synthetic images generated (a) by CIFAR10 generator $G_p$ through MI with SVHN; (b) by $G_p$ through GOPMI with SVHN.}
\label{fig:gmi_vs_mi}
\end{figure}

In this study, we first compare the identified private latent vector $z_s$ from GOMI and MI using FID and IS values. As shown in Table~\ref{tab:cap}, GOMI has 4\% to 10\% smaller values of FID and 4\% to 5\% higher IS scores, which indicates a better performance in terms of inversion quality. We further stress test the two schemes by using the entire CIFAR10 as $D_p$ and SVHN as $D_s$. A visual comparison between MI and GOMI is shown in Figure~\ref{fig:gmi_vs_mi}. Both MI and GOMI can perform inversion using a completely different dataset, and the quality of inverted images from GOMI is better than the ones from MI.
\begin{table}[h!]
\begin{center}
\ra{0.9}
\caption{FID and IS between MI and GOMI} \label{tab:cap}
\begin{tabular}{@{}cccc@{}}
  \toprule
  Metric & Dataset & $G_p$ + MI~\cite{chen2021differentially} & $G_p$ + GOMI
  \\
  \midrule
  FID & CIFAR10 & 81.74 & \bf{78.57} \\
      & SVHN & 26.11 & \bf{23.40} \\
  IS  & CIFAR10 & 4.61 & \bf{4.83} \\
      & SVHN & 2.72 & \bf{2.82} \\
  \bottomrule
\end{tabular}
\end{center}
\end{table}

\section{Conclusion}

In this work, we have proposed a new differentially private data publishing method, referred as DPGOMI. The proposed Gaussian Optimized Model Inversion method improves upon the existing model inversion process by providing a differential cost function that favors the target latent space distribution. Using two public datasets, the proposed DPGOMI have demonstrated superior performance both quantitatively and qualitatively in the comparison with SOTA methods and in the ablation study. 

\bibliographystyle{IEEEtran}
\bibliography{ICIPW2023}

\begin{thebibliography}{10}
\providecommand{\url}[1]{#1}
\csname url@samestyle\endcsname
\providecommand{\newblock}{\relax}
\providecommand{\bibinfo}[2]{#2}
\providecommand{\BIBentrySTDinterwordspacing}{\spaceskip=0pt\relax}
\providecommand{\BIBentryALTinterwordstretchfactor}{4}
\providecommand{\BIBentryALTinterwordspacing}{\spaceskip=\fontdimen2\font plus
\BIBentryALTinterwordstretchfactor\fontdimen3\font minus
  \fontdimen4\font\relax}
\providecommand{\BIBforeignlanguage}[2]{{%
\expandafter\ifx\csname l@#1\endcsname\relax
\typeout{** WARNING: IEEEtran.bst: No hyphenation pattern has been}%
\typeout{** loaded for the language `#1'. Using the pattern for}%
\typeout{** the default language instead.}%
\else
\language=\csname l@#1\endcsname
\fi
#2}}
\providecommand{\BIBdecl}{\relax}
\BIBdecl

\bibitem{abadi2016deep}
M.~Abadi, A.~Chu, I.~Goodfellow, H.~B. McMahan, I.~Mironov, K.~Talwar, and
  L.~Zhang, ``Deep learning with differential privacy,'' in \emph{Proceedings
  of the 2016 ACM SIGSAC Conference on Computer and Communications Security},
  2016, pp. 308--318.

\bibitem{bu2021convergence}
Z.~Bu, H.~Wang, Q.~Long, and W.~J. Su, ``On the convergence of deep learning
  with differential privacy,'' \emph{arXiv e-prints}, pp. arXiv--2106, 2021.

\bibitem{chen2021differentially}
D.~Chen, S.-c.~S. Cheung, C.-N. Chuah, and S.~Ozonoff, ``Differentially private
  generative adversarial networks with model inversion,'' in \emph{2021 IEEE
  International Workshop on Information Forensics and Security (WIFS)}.\hskip
  1em plus 0.5em minus 0.4em\relax IEEE, 2021, pp. 1--6.

\bibitem{krizhevsky2009learning}
A.~Krizhevsky, G.~Hinton \emph{et~al.}, ``Learning multiple layers of features
  from tiny images,'' 2009.

\bibitem{netzer2011reading}
Y.~Netzer, T.~Wang, A.~Coates, A.~Bissacco, B.~Wu, and A.~Y. Ng, ``Reading
  digits in natural images with unsupervised feature learning,'' 2011.

\bibitem{dwork2014algorithmic}
C.~Dwork, A.~Roth \emph{et~al.}, ``The algorithmic foundations of differential
  privacy.'' \emph{Foundations and Trends in Theoretical Computer Science},
  vol.~9, no. 3-4, pp. 211--407, 2014.

\bibitem{papernotscalable}
N.~Papernot, S.~Song, I.~Mironov, A.~Raghunathan, K.~Talwar, and U.~Erlingsson,
  ``Scalable {Private} {Learning} with {PATE},'' in \emph{International
  Conference on Learning Representations}, 2018.

\bibitem{harder2023pre}
F.~Harder, M.~Jalali, D.~J. Sutherland, and M.~Park, ``Pre-trained perceptual
  features improve differentially private image generation,''
  \emph{Transactions on Machine Learning Research}, 2023.

\bibitem{fan2020survey}
L.~Fan, ``A survey of differentially private generative adversarial networks,''
  in \emph{The AAAI Workshop on Privacy-Preserving Artificial Intelligence},
  2020.

\bibitem{chen2020gs}
D.~Chen, T.~Orekondy, and M.~Fritz, ``{GS-WGAN}: A gradient-sanitized approach
  for learning differentially private generators,'' \emph{Advances in Neural
  Information Processing Systems}, vol.~33, 2020.

\bibitem{jordon2018pate}
J.~Jordon, J.~Yoon, and M.~van~der Schaar, ``{PATE-GAN}: Generating synthetic
  data with differential privacy guarantees,'' in \emph{International
  Conference on Learning Representations}, 2018.

\bibitem{long2021g}
Y.~Long, B.~Wang, Z.~Yang, B.~Kailkhura, A.~Zhang, C.~Gunter, and B.~Li,
  ``{G-PATE}: scalable differentially private data generator via private
  aggregation of teacher discriminators,'' \emph{Advances in Neural Information
  Processing Systems}, vol.~34, pp. 2965--2977, 2021.

\bibitem{santos2019learning}
C.~N.~d. Santos, Y.~Mroueh, I.~Padhi, and P.~Dognin, ``Learning implicit
  generative models by matching perceptual features,'' in \emph{Proceedings of
  the IEEE/CVF International Conference on Computer Vision}, 2019, pp.
  4461--4470.

\bibitem{harder2021dp}
F.~Harder, K.~Adamczewski, and M.~Park, ``{DP-MERF}: Differentially private
  mean embeddings with randomfeatures for practical privacy-preserving data
  generation,'' in \emph{International conference on artificial intelligence
  and statistics}.\hskip 1em plus 0.5em minus 0.4em\relax PMLR, 2021, pp.
  1819--1827.

\bibitem{mironov2017renyi}
I.~Mironov, ``R{\'e}nyi differential privacy,'' in \emph{2017 IEEE 30th
  Computer Security Foundations Symposium (CSF)}.\hskip 1em plus 0.5em minus
  0.4em\relax IEEE, 2017, pp. 263--275.

\bibitem{arjovsky2017wasserstein}
M.~Arjovsky, S.~Chintala, and L.~Bottou, ``Wasserstein {GAN},'' \emph{arXiv
  preprint arXiv:1701.07875}, 2017.

\bibitem{shalev2014understanding}
S.~Shalev-Shwartz and S.~Ben-David, \emph{Understanding machine learning: From
  theory to algorithms}.\hskip 1em plus 0.5em minus 0.4em\relax Cambridge
  university press, 2014.

\bibitem{marchesi2017megapixel}
M.~Marchesi, ``Megapixel size image creation using generative adversarial
  networks,'' \emph{arXiv preprint arXiv:1706.00082}, 2017.

\bibitem{xie2018differentially}
L.~Xie, K.~Lin, S.~Wang, F.~Wang, and J.~Zhou, ``Differentially private
  generative adversarial network,'' \emph{arXiv preprint arXiv:1802.06739},
  2018.

\bibitem{borji2021pros}
A.~Borji, ``Pros and {Cons} of {GAN} evaluation measures: New developments,''
  \emph{arXiv preprint arXiv:2103.09396}, 2021.

\end{thebibliography}

\end{document}